\documentclass[conference]{IEEEtran}
\IEEEoverridecommandlockouts
\usepackage{cite}
\usepackage{amsmath,amssymb,amsfonts}
\usepackage{algorithmic}
\usepackage{graphicx}
\usepackage{subfig}
\usepackage{textcomp}
\usepackage{xcolor}
\usepackage{url}
\def\BibTeX{{\rm B\kern-.05em{\sc i\kern-.025em b}\kern-.08em
    T\kern-.1667em\lower.7ex\hbox{E}\kern-.125emX}}
\begin{document}

\title{Scalable Power Control/Beamforming in Heterogeneous Wireless Networks with Graph Neural Networks\\
\thanks{This work was supported in part by the National Natural Science Foundation of China under Grants 61931020, U19B2024, 61801494 and 62101569.}
}

\author{\IEEEauthorblockN{Xiaochen Zhang, Haitao Zhao, Jun Xiong, Xiaoran Liu, Li Zhou, Jibo Wei}
\IEEEauthorblockA{\textit{College of Electronic Science and Technology} \\
National University of Defense Technology, Changsha, China, 410073 \\
\{zhangxiaochen14, haitaozhao, xj8765, liuxiaoran10, zhouli2035, wjbhw\}@nudt.edu.cn}
}

\maketitle

\begin{abstract}
Machine learning (ML) has been widely used for efficient resource allocation (RA) in wireless networks. Although superb performance is achieved on small and simple networks, most existing ML-based approaches are confronted with difficulties when heterogeneity occurs and network size expands. In this paper, specifically focusing on power control/beamforming (PC/BF) in heterogeneous device-to-device (D2D) networks, we propose a novel unsupervised learning-based framework named heterogeneous interference graph neural network (HIGNN) to handle these challenges. First, we characterize diversified link features and interference relations with heterogeneous graphs. Then, HIGNN is proposed to empower each link to obtain its individual transmission scheme after limited information exchange with neighboring links. It is noteworthy that HIGNN is scalable to wireless networks of growing sizes with robust performance after trained on small-sized networks. Numerical results show that compared with state-of-the-art benchmarks, HIGNN achieves much higher execution efficiency while providing strong performance.
\end{abstract}


\section{Introduction}
Given the time-varying characteristics of wireless fading channels, economical allocation of limited power budget and spectrum plays a central role in the performance enhancement of wireless communications networks (WCNs). Facing the ever-increasing number of users with diverse requirements on service, a variety of utility functions such as weighted sum rate (WSR) and energy efficiency (EE) are developed to assess the system performance in different scenarios. Unfortunately, maximizing these metrics under practical constraints in general interference channels (IFCs) is scarcely a trivial issue. Due to mutual interference among links and strong coupling of optimization variables, many of these problems are non-convex and solving them has been already proved to be NP-hard. 

A plethora of research has been carried out to find efficient optimizers and a most well-known candidate is the weighted minimum mean squared error (WMMSE) minimization \cite{Shi:2011tv} for WSR maximization. A globally optimal solution is found in \cite{Bjornson:2013us} at the cost of computational complexity. Later, K. Shen and W. Yu \cite{Shen:2018th} extend the fractional programming (FP) theory for efficient suboptimal solutions to general resource allocation (RA) in WCNs. Although the theoretical throughputs of WCNs have been greatly elevated by aforementioned algorithms, many obstacles are placed in their way towards practical implementation. Firstly, these algorithms are iterative and complicated computation (including matrix inversion) is involved in each iteration. Secondly, many of these algorithms are executed in a centralized manner, resulting in enormous overheads on control links. 

In recent years, machine learning (ML) is shown to be a competitive candidate for solving non-convex RA problems. A `learn-to-optimize' approach \cite{Sun:2018vn} is initially put forward by H. Sun, \textit{et al.}, where deep neural networks (DNNs) are adopted to imitate the input-output mapping of WMMSE in a supervised manner. Pre-trained neural network (NN) models require much less computation than WMMSE to achieve matching performance. The main drawbacks of this approach lie in the computational burden of label generation from WMMSE and the poor generalization ability to large problem scales. Alternatively, unsupervised learning is used in \cite{Lee:2018wq, Eisen:2019ti, Liang:2020uz} to acquire RA policies by directly maximizing utility functions. To enable trained NNs to handle wireless network of different sizes, a convolution-based method is proposed in \cite{Cui:2019wh} for link scheduling. However, topology instead of direct channel state information (CSI) is utilized in \cite{Cui:2019wh}, thereby curtailing the performance under fading channels. 

Interference relations in WCNs can be naturally described by graphs, and in consequence graph neural networks (GNNs) can be used for RA in WCNs. The major advantage of GNNs is their transference ability to different sizes of WCNs, which pragmatically takes into account the fluctuation in the number of active links over time. Based on \cite{Eisen:2019ti}, M. Eisen and A. Ribeiro propose the random edge graph neural network (REGNN) \cite{Eisen:2020aa} for power control (PC). Graph embedding is adopted in \cite{Lee:2020aa} for link scheduling without leveraging accurate CSI. \cite{zhao2021distributed} applies graph convolutional network (GCN) to the scheduling of orthogonal resource blocks after modeling it as a maximum weighted independent set (MWIS) problem. Then Y. Shen \textit{et al.} propose distributed algorithms to solve a variety of optimization problems including PC, beamforming (BF) \cite{Shen:2021wb} and statistical inference \cite{wang2021decentralized} based on GNN. Further, in \cite{chowdhury2021unfolding}, A. Chowdhury \textit{et al.} modifies WMMSE algorithm by inserting a learnable GNN module to facilitate convergence. 

Nonetheless, aforementioned ML-based methods share the homogeneity assumption on WCNs without exception. Even GNN-based methods could only tackle arbitrary numbers of links with the same properties. By contrast, modern networks inevitably encompass multiple types of links with disparate properties. In this case, traditional methods like WMMSE still work by executing update for each link type, while most ML-based counterparts would fail. As a result, we are interested in investigating how to extend the scalability of GNNs to RA problems in heterogeneous settings. We first characterize the interference relations with heterographs. Then, we propose a framework called heterogeneous interference graph neural network (HIGNN) to solve RA problems for wireless links with diversified features. Most recently, heterogeneous GNNs are used by \cite{Guo:2021vb} to learn PC in a supervised manner, for which the motive is to separately treat transmitters and receivers as different types of vertices. In contrast, we define vertices as communication links and pay attention to manage the complex interference relations together with incompatible action space of multiple link types. Moreover, our model is based on unsupervised learning. This work could be considered as a generalization of \cite{Eisen:2020aa, Shen:2021wb} from homogeneous networks to more complex heterogeneous systems. 

\begin{figure*}[!t]
\centering
\subfloat[An example of a heterogeneous D2D network with two types of links. When links have different features, so does the interference they cause.]{\includegraphics[width=3in]{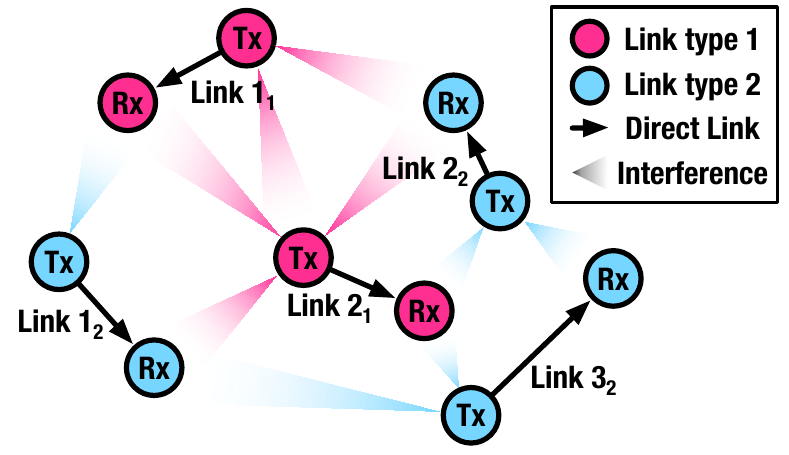}%
\label{fig_first_case}}
\hfil
\subfloat[A heterograph describing the interference pattern of the heterogeneous D2D network in (a).]{\includegraphics[width=2.8in]{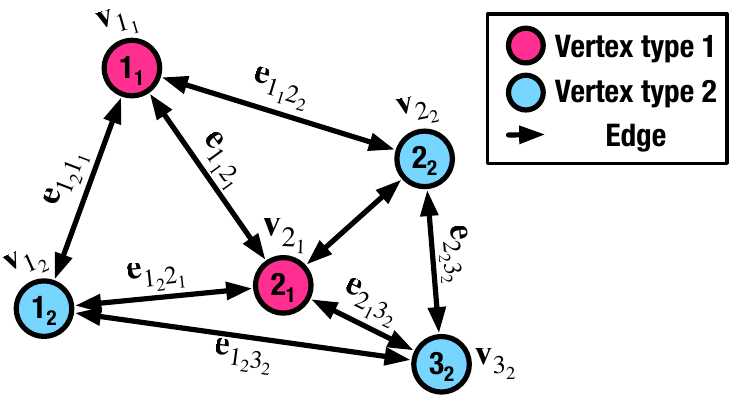}%
\label{fig_second_case}}
\caption{Modeling of heterogeneous IFCs as heterographs. Each link is treated as a vertex and interference relations define the edges among vertices. Link index $i_{m}$ means the $i$-th link of type $m$.}
\label{fig_sim}
\end{figure*}

\section{System Model and Problem Formulation}
A single-hop D2D network is considered, where multiple transceiver pairs share the same spectrum of bandwidth $W$. In heterogeneous settings, different links are permitted to hold varied features. Here, we assume that each receiver is equipped with single antenna while the number of transmit antennas may differ from one link to another. Denote the $M$ types of links by $\mathcal{M}\triangleq\{1, \cdots, M\}$, and the number of transmit antennas for link type $m$ is $N_{m}$. The number of links belonging to type $m$ is $K_{m}$ and the total number of links is $K=\sum_{m}K_{m}$. Index the $i$-th link of type $m$ as $i_{m}$, and then channel response from the transmitter of link $j_{n}$ to the receiver of link $i_{m}$ is $\mathbf{h}_{i_{m}j_{n}}\in\mathbb{C}^{N_{n}}$. Let $\mathbf{x}_{i_{m}}\in\mathbb{C}^{N_{m}}$ be the beamforming vector of link $i_{m}$. Accordingly, the received signal at receiver $i_{m}$ is the superposition of signals from multiple types of transmitters, which is given by
\begin{equation}
y_{i_{m}} = \mathbf{h}_{i_{m}i_{m}}^{H}\mathbf{x}_{i_{m}}s_{i_{m}} + \sum_{ j_{n}\not=i_{m}}\mathbf{h}_{i_{m}j_{n}}^{H}\mathbf{x}_{j_{n}}s_{j_{n}} + n_{i_{m}},
\end{equation}
where $s_{i_{m}}\sim\mathcal{CN}(0, 1)$ denotes the transmitted symbol and $n_{i_{m}}\sim\mathcal{N}(0, \sigma_{i_{m}}^{2})$ represents the additive white Gaussian noise (AWGN) with zero mean and variance $\sigma_{i_{m}}^{2}$. The achievable rate of link $i_{m}$ is a function of beamformer $\mathbf{X}\triangleq\{\mathbf{x}_{i_{m}}\}_{i_{m}}$: 
\begin{equation}
R_{i_{m}}(\mathbf{X}) = W\log\Bigg(1 + \frac{\parallel\!\mathbf{h}_{i_{m}i_{m}}^{H}\mathbf{x}_{i_{m}}\!\parallel_{2}^{2}}{\sum_{j_{n}\not=i_{m}}\parallel\!\mathbf{h}_{i_{m}j_{n}}^{H}\mathbf{x}_{j_{n}}\!\parallel_{2}^{2} + \sigma_{i_{m}}^{2}}\Bigg).
\end{equation}
Particularly, beamforming design reduces to PC when single antenna is used at transmitters, \textit{i.e.}, $N_{m}=1$.

Overall performance of the network is usually evaluated by a utility function of achievable rates for all links. Here, we choose WSR. Given the power constraint at each transmitter, the optimization problem is formulated as 
\begin{equation}
\label{problem}
\begin{aligned}
\max_{\mathbf{X}} & \sum_{i, m}\omega_{i_{m}}R_{i_{m}}(\mathbf{X})\\
{\rm s.t.} & \parallel\!\mathbf{x}_{i_{m}}\!\!\parallel_{2}^{2}\leq P_{\rm max}, \forall i, m,
\end{aligned}
\end{equation}
with weight $\omega_{i_{m}}$ indicating the priority of link $i_{m}$. When all weights are set to 1, problem described by Eq. (\ref{problem}) turns to sum rate maximization.

\section{Scalable PC/BF in Heterogeneous D2D Networks with HIGNN}
In this section, we show how to describe the interference relations between different types of links using heterogeneous graphs and then propose an efficient learning framework for utility-maximizing PC/BF in heterogeneous settings. 

\subsection{Relational Modeling of Heterogeneous IFCs}
A graph can be formally characterized by a tuple $\mathcal{G} = (\mathcal{V}, \mathcal{E})$, where $\mathcal{V}$ and $\mathcal{E}$ are sets containing vertices and edges, respectively. A vertex $i\in\mathcal{V}$ represents an entity and an edge $(i, j)\in\mathcal{E}$ defines a directed relation from vertex $i$ to vertex $j$. Denote the neighboring set of vertex $i$ as $\mathcal{N}_{i}=\{j\in\mathcal{V}|(j, i)\in\mathcal{E}\}$. Attributes of vertex $i$ and edge $(i, j)$ are characterized by $\mathbf{v}_{i}$ and $\mathbf{e}_{ij}$, respectively.

When multiple types of vertices or edges occur, the graph becomes \textit{heterogeneous} \cite{Schlichtkrull:2018vh}. Define the set of vertex types as $\mathcal{T}$ and relation types as $\mathcal{R}$. Here, relations are adopted to identify vertex types associated with edges. Denote the $i$-th vertex of type $m$ by $i_{m}$ and refer to its incident neighbor vertices under relation $r=(n, m)$ by $\mathcal{N}_{i_{m}}^{(n)}=\{j|(j_{n}, i_{m})\in\mathcal{E}\}$. For notational simplicity, vertex attributes are held by $\mathbf{V}\triangleq\{\mathbf{V}_{m}\}_{m}$, where $m$ specifies vertex types and $[\mathbf{V}_{m}]_{i}=\mathbf{v}_{i_{m}}$. Edge attributes are collectively given as $\mathbf{E}\triangleq\{\mathbf{E}_{mn}\}_{m, n}$, where $[\mathbf{E}_{mn}]_{ij} = \mathbf{e}_{i_{m}j_{n}}$ if edge $(i_{m}, j_{n})$ exists and $\mathbf{0}$ otherwise. 

In the context of wireless interference, we treat each transceiver pair as a vertex and the interference pattern from transmitters to receivers as edges. Attributes of each vertex $i_{m}$ may include the weight $\omega_{i_{m}}$, noise variance $\sigma_{i_{m}}^{2}$ and direct channel response $\mathbf{h}_{i_{m}i_{m}}$. Each edge $(i_{m}, j_{n})$ is characterized by channel response from the interfering transmitters to the interfered receivers, \textit{i.e.}, $\mathbf{e}_{i_{m}j_{n}}=[\mathbf{h}_{i_{m}j_{n}}, \mathbf{h}_{j_{n}i_{m}}]$. As different link types are equipped with different numbers of Tx antennas, dimensions of vertex/edge attributes may vary with types.

\subsection{Sum-Rate Maximization via Convolution on Heterogeneous Interference Graphs }
With interference relations modeled as a heterograph $G$ of channel coefficients $\mathbf{H}\triangleq\{\mathbf{h}_{i_{m}j_{n}}\}_{i_{m}, j_{n}}$, the goal of interest is to find a policy $p(\cdot)$ mapping the heterograph to estimates of optimal beamforming vectors $\hat{\mathbf{X}}\triangleq\{\hat{\mathbf{x}}_{i_{m}}\}_{i_{m}}$. We choose a GNN parameterization of policy $p_{\boldsymbol{\theta}}(\cdot)$ with learnable parameters $\boldsymbol{\theta}$, and beamforming vectors are estimated as $\hat{\mathbf{X}}=p_{\boldsymbol{\theta}}(G)$. We refer to the proposed framework as heterogeneous interference graph neural network (HIGNN). Its principles and compositions are explained in the following. 

\subsubsection{Graph Convolution}
The basic computation unit over graphs is \textit{graph network (GN) block} \cite{battaglia2018relational}, which includes \textit{update functions} $\phi$ and \textit{aggregation functions} $\rho$. In our case, GN blocks need to yield outputs at vertices as transmission scheme for each link. Accordingly, update functions and aggregation functions are defined as
\begin{equation}
\label{update_funcs}
{\rm Update}:
\begin{cases}
\mathbf{e}_{ij}[l] = \phi^{e}(\mathbf{v}_{i}[l-1], \mathbf{e}_{ij}[l-1], \mathbf{v}_{j}[l-1]),\\
\mathbf{v}_{i}[l] = \phi^{v}(\bar{\mathbf{e}}_{i}[l], \mathbf{v}_{i}[l-1]),\\
\end{cases}
\end{equation}
\begin{equation}
{\rm Aggregation}: \bar{\mathbf{e}}_{i}[l] = \rho^{e\rightarrow v}(\{\mathbf{e}_{ji}[l]\}_{j\in\mathcal{N}_{i}}),\\
\end{equation}
with $l$ indexing the  current step of update. $\phi^{e}$ is first applied to each edge to encode vertex/edge attributes. Then each vertex $i$ aggregates the updates of edges $(j, i)$, $\forall j\in\mathcal{N}_{i}$, with $\rho^{e\rightarrow v}$. Later, $\phi^{v}$ is employed to obtain the vertex update by combining the aggregated edge update $\bar{\mathbf{e}}_{i}[l]$ and its current attributes $\mathbf{v}_{i}[l-1]$. Message-passing is completed as knowledge at each vertex is embedded in edge updates and subsequently absorbed by its neighboring vertices. Common choices of update functions are NN modules. Summation, mean and max/min are usually taken as aggregation functions. The update at each vertex takes place independently and thus the implementation of GNN-based RA algorithms is regarded as distributed.

\subsubsection{Design of GN Blocks for Heterogeneous PC/BF}

In heterographs, properties of attributes may vary from one vertex/edge type to another. For instance, in heterogeneous IFCs described in Section II, vertex/edge attributes involve channel response, of which dimensions change with antenna numbers. These features from different relations cannot be handled by plain GNNs described above and they must be treated separately. As suggested in \cite{han2019}, each relation is assigned individual aggregation/update functions. Message-passing is first executed within each relation. Later, destination vertices sample and aggregate partial updates from multiple relations to obtain their final updates.

\begin{figure}[htbp]
\centerline{\includegraphics[width=3.5in]{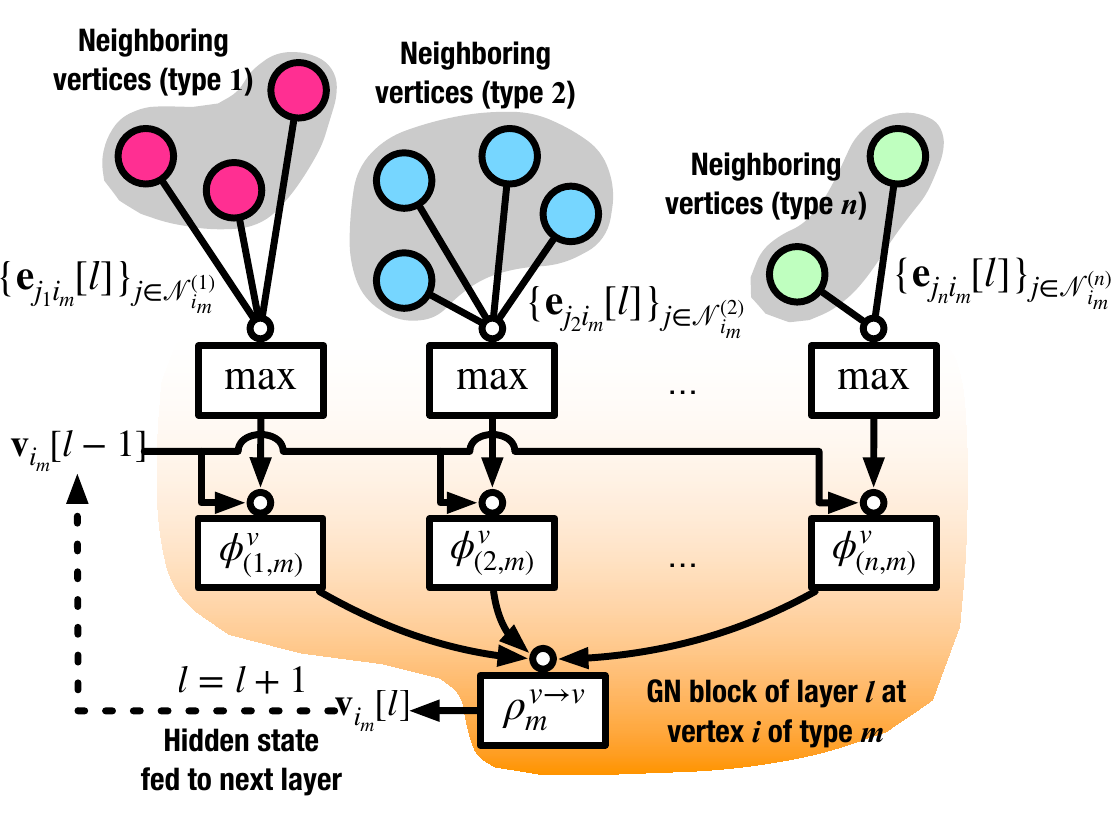}}
\caption{Diagram of heterograph convolution at vertex $i_{m}$. An individual vertex update function $\phi_{(n, m)}^{v}$ is defined for each relation $(n, m)$ to produce a partial update $\mathbf{v}_{i_{m}}^{(n)}[l]$. Final update $\mathbf{v}_{i_{m}}[l]$ is the combination of partial updates across relations.}
\label{hetero_gn_blk}
\end{figure}

Following the above principles, we turn to specific design of GN blocks for heterogeneous PC/BF. Relations in our context are distinguished by the types of interfering/interfered links according to their numbers of transmit antennas. Relation $r=(n, m)$ characterizes the interference from link type $n$ to link type $m$. For each relation $r$, we define an edge update function $\phi_{r}^{e}$ and a vertex update function $\phi_{r}^{v}$ which are parametrized by multi-layer perceptrons (MLPs). Without loss of generality, define the partial update at vertex $i_{m}$ in relation $(n, m)$ as:
\begin{equation}
\begin{aligned}
\mathbf{e}_{j_{n}i_{m}}[l]&= \phi_{(n, m)}^{e}(\mathbf{v}_{j_{n}}[l-1], \mathbf{e}_{j_{n}i_{m}}[0]) ,\\
\mathbf{v}_{i_{m}}^{(n)}[l] &= \phi_{(n, m)}^{v}(\mathbf{v}_{i_{m}}[l-1], \max_{j\in\mathcal{N}_{i_{m}}^{(n)}}\mathbf{e}_{j_{n}i_{m}}[l]).
\end{aligned}
\end{equation}
Note that several modifications from general guideline Eq. (\ref{update_funcs}) are made owing to practical concerns about distributed implementation: 1) The attributes of destination vertices in $\phi_{(n, m)}^{e}$ are dropped for reduction in information exchange; 2) Initial edge attributes $\mathbf{e}_{j_{n}i_{m}}[0]$ are kept used in all steps of edge update. We also empirically find that for $l>1$, using the concatenation of the last vertex update and initial vertex attributes $[\mathbf{v}_{i_{m}}[l-1], \mathbf{v}_{i_{m}}[0]]$ as input to $\phi_{(n, m)}^{v}$ helps to stabilize training performance. Then per-relation updates $\{\mathbf{v}_{i_{m}}^{(n)}[l]\}_{n}$ are merged to get the final vertex update $\mathbf{v}_{i_{m}}[l]$ as
\begin{equation}
\mathbf{v}_{i_{m}}[l] = \rho_{m}^{v\rightarrow v}(\{\mathbf{v}_{i_{m}}^{(n)}[l]\}_{n}) =\frac{1}{c_{i, m}}\sum_{n} \mathbf{v}_{i_{m}}^{(n)}[l],
\end{equation}
where $c_{i, m}$ is the number of relations involved in updating $\mathbf{v}_{i_{m}}[l]$. A diagram for these procedures is given in Fig. \ref{hetero_gn_blk}. Specially, when all communication links belong to the same type as assumed in \cite{Eisen:2020aa, Shen:2021wb}, heterogeneous interference graphs reduce to homogeneous and plain graph convolution occurs as a special case of our model. 

\subsubsection{Model Architecture of HIGNN}

Suppose that $L$ steps of updates are executed in total, and hence $L$ concatenated convolution layers (GN blocks) are defined as ${\rm GN}_{1},\dots,{\rm GN}_{L}$. To facilitate training, we adopt an encode-process-decode \cite{battaglia2018relational} architecture which consists of an encoder ${\rm GN}_{\rm enc}$, a shared core block ${\rm GN}_{\rm core}$ and a decoder ${\rm GN}_{\rm dec}$. Here, all intermediate convolution layers share the same parameters, \textit{i.e.}, ${\rm GN}_{2}=\cdots={\rm GN}_{L-1}={\rm GN}_{\rm core}$ when $L\geq3$. 

During forward computation, each vertex $i_{m}$ takes its attributes as the initial input $\mathbf{v}_{i_{m}}[0]$ to ${\rm GN}_{\rm enc}$ and yields the embedding $\mathbf{v}_{i_{m}}[1]$. Then, ${\rm GN}_{\rm core}$ recurrently produces update $\mathbf{v}_{i_{m}}[l]$ from $\mathbf{v}_{i_{m}}[l-1]$ for $l<L$. Eventually, $\mathbf{v}_{i_{m}}[L-1]$ is passed to ${\rm GN}_{\rm dec}$ to obtain the estimate of beamforming vector $\hat{\mathbf{x}}_{i_{m}}=\mathbf{v}_{i_{m}}[L]$. At the output of ${\rm GN}_{\rm dec}$, power constraint is imposed by activation function $\gamma(\mathbf{x})=\frac{\sqrt{P_{\max}}\mathbf{x}}{\max\{\parallel\mathbf{x}\parallel_{2}, 1\}}$ . 

The loss function $\mathcal{L}$ is the negative expectation of utility function over different channel realizations:
\begin{equation}
\label{loss_func}
\mathcal{L}(\boldsymbol{\theta})\!=\!-\mathsf{E}_{\mathbf{H}}\Bigg[\!\sum_{i, m}\omega_{i_{m}}\!\log\!\Bigg(\!1 + \frac{\parallel\!\mathbf{h}_{i_{m}i_{m}}^{H}\hat{\mathbf{x}}_{i_{m}}\!\!\parallel_{2}^{2}}{\sum_{j_{n}\not=i_{m}}\!\!\parallel\!\!\mathbf{h}_{i_{m}j_{n}}^{H}\hat{\mathbf{x}}_{j_{n}}\!\!\parallel_{2}^{2} + \sigma_{i_{m}}^{2}}\!\Bigg)\!\Bigg].
\end{equation}
Backpropagation on Eq. (\ref{loss_func}) is done to update model parameters of HIGNN $\boldsymbol{\theta}$ in an unsupervised fashion. Theoretically, stacking more convolution layers tend to result in better performance while bringing more burden on computation and information exchange among vertices. A trade-off can be found via extensive experiments on specific datasets. 

\subsection{Permutation Invariance and Equivalence Properties}
Two fundamental properties of GNNs are permutation invariance and permutation equivalence, which underpin the transference ability of GNNs to different problem instances. Permutation invariance suggests that permutation of vertices is independent of the output, while permutation equivalence ensures the permutation of inputs leads to the same permutation at outputs. While these two properties are addressed in \cite{Eisen:2020aa, Shen:2021wb} for homogeneous settings, here we extend the analysis to heterogeneous cases.

Permutation matrices are defined as $\mathbf{\Pi}=\{\mathbf{\Pi}_{m}\}_{m}$ where $\mathbf{\Pi}_{m}\in\{\{0, 1\}^{K_{m}\times K_{m}}|, \boldsymbol{\Pi}_{m}^{T}\mathbf{1}=\mathbf{1}\boldsymbol{\Pi}_{m}=\mathbf{1}\}$. Multiplication of a matrix by the permutation matrix leads to the reordering of the rows/columns of the matrix. Then we define a composite version of permutation over heterogeneous graphs: $\mathbf{\Pi}\mathbf{V} = \{\mathbf{\Pi}_{m}\mathbf{V}_{m}\}_{m}$ and $\mathbf{\Pi}\mathbf{E}\mathbf{\Pi}=\{\mathbf{\Pi}_{m}\mathbf{E}_{mn}\mathbf{\Pi}_{n}\}_{m, n}$. The invariance of utility function to the permutation is given by
\begin{equation}
f(\mathbf{\Pi X}, \mathbf{\Pi V}, \mathbf{\Pi}^{T}\mathbf{E}\mathbf{\Pi}) = f(\mathbf{X}, \mathbf{V}, \mathbf{E}),
\end{equation}
while the permutation equivalence property of the policy $p_{\boldsymbol{\theta}}$ by GNN is expressed as
\begin{equation}
\mathbf{\Pi}\hat{\mathbf{X}}=p_{\boldsymbol{\theta}}(\mathbf{\Pi V}, \mathbf{\Pi}^{T}\mathbf{E}\mathbf{\Pi}).
\end{equation}
These properties suggest that the reordering of vertices (links) does not affect the outputs of GNNs (beamforming schemes) at each vertex, thereby leaving the global utility function remain the same. Compared with other models such as DNNs and convolutional neural networks (CNNs), GNNs can learn RA policies that are more consistent in different wireless networks.

\section{Simulation Results and Analysis}
In this section, the setting of simulation and model details are introduced, followed by experimental results and corresponding analysis. For those who are interested in our work, we provide an implementation of HIGNN using deep graph library (DGL) \cite{wang2019dgl} with PyTorch backend\footnote{Code is available at \url{https://github.com/zhangxiaochen95/hignn}.}.

\subsection{Simulation Setup and Model Specifications}
We simulate a D2D network where all links share the same bandwidth. Two types of links are considered: 1) SISO links and 2) $2\times1$ MISO links. All transmitters and receivers are uniformly located in a square area of length $D$. The communication range of each link is restricted between $d_{\min} = 2{\rm m}$ and $d_{\max} = 50{\rm m}$. Channel response is computed by $\mathbf{h}_{i_{m}j_{n}}=\sqrt{\beta_{i_{m}j_{n}}}\mathbf{g}_{i_{m}j_{n}}$, where $\beta_{i_{m}j_{n}}\in\mathbb{R}$ and $\mathbf{g}_{i_{m}j_{n}}\in\mathbb{C}^{N_{n}}$ stand for large-scale fading and small-scale fading components, respectively. Small-scale fading is represented by i.i.d. zero-meaned complex Gaussian variables with unit variance. A scaled distance-dependent model adopted by \cite{Shen:2021wb} is used to determine path loss and shadowing in large-scale fading. Similar to \cite{Sun:2018vn, Eisen:2020aa}, noise variance at receiver and transmit power budget are normalized to $1$. During the generation of channel instances for training, numbers of SISO and MISO links are set to 8 and 4, respectively. The length of area is $D=400{\rm m}$ and network topology for each sample is determined independently. 

Since channel responses are complex vectors, real part and imaginary part of $\mathbf{H}$ are separately fed to NN modules after normalization. In this manner, sizes of input features for vertex type $m$ and relation $r=(n, m)$ are $2N_{m}$ and $2N_{n} + 2N_{n}$, respectively. Sizes of all messages and intermediate updates are set to 8. Dimension of output features  for vertex type $m$ is $2N_{m}$. $\phi_{(n, m)}^{e}$ and $\phi_{(n, m)}^{v}$ for all $n, m$ in all GN blocks are parameterized by MLPs of hidden size $\{16\}$ unless otherwise specified. Adam optimizer \cite{kingma2017adam} is used with learning rate set to $0.001$. 

For performance evaluation of HIGNN, we mainly compare it with the FP algorithm in \cite{Shen:2018th}. In sum rate maximization, the closed-form FP is shown to be equivalent to WMMSE, a widely used benchmark in literature. All results on test performance are the average from 1000 independent trials. 

\subsection{Training Efficiency}
\begin{figure}[htbp]
\centerline{\includegraphics[width=3.5in]{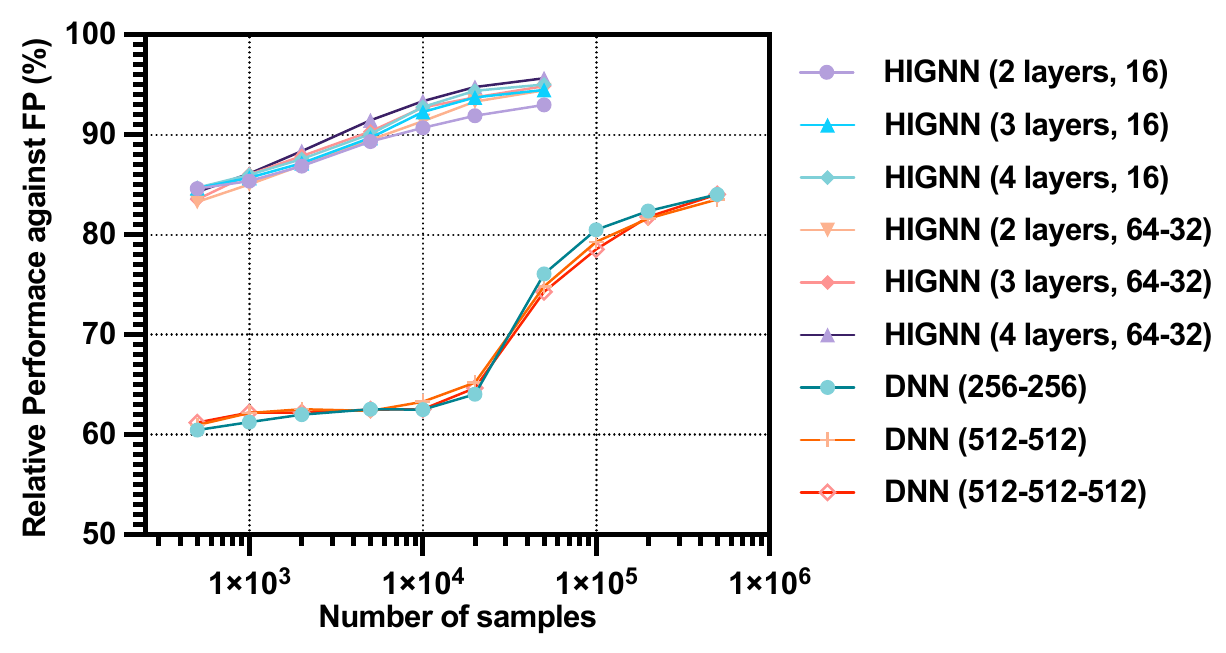}}
\caption{Relative performance of models against FP with respect to size of training set (in logarithm). Number of convolution layers and sizes of hidden layers in update functions are shown in the bracket for HIGNNs. The figures for DNNs of different sizes are also presented to compare with HIGNN.}
\label{hyperparameters}
\end{figure}

Fig. \ref{hyperparameters} illustrates the relative performance of HIGNNs against FP with respect to the size of training set. A two-layer HIGNN (without ${\rm GN}_{\rm core}$) with MLP hidden size $\{16\}$ achieves approximately $85\%$ the performance of FP when trained on mere 500 samples. The performance of HIGNN could be enhanced by either adding more convolution layers or using larger MLPs to parameterize $\phi_{(n, m)}^{e}$ and $\phi_{(n, m)}^{v}$. Particularly, we give the results when hidden layer size of MLPs in $\phi_{(n, m)}^{e}$ and $\phi_{(n, m)}^{v}$ is $\{64, 32\}$ to demonstrate the benefits of enlarging MLPs. The relative performance is raised to around $96\%$ by a 4-layer HIGNN with MLP hidden size $\{64, 32\}$. We use a 3-layer HIGNN with MLP hidden size $\{16\}$ (with relative performance above $95\%$) in the following experiments for trade-off between performance and complexity. 

Here, we also present the results of DNNs to illustrate the remarkable training efficiency of HIGNN. The input dimension of DNNs is $\sum_{n}\sum_{m}2K_{n}\cdot K_{m}\cdot N_{n}$ and output dimension is $\sum_{m}2K_{m}\cdot N_{m}$. Compared with HIGNN, DNNs generally require much more samples and still show much worse performance. The relative performance of a DNN with hidden size $\{512,512\}$ is only $83.5\%$ when 500,000 samples are fed. Adding additional hidden layers to DNNs does not make a distinct improvement and conversely it is liable to cause overfitting. Overall, HIGNNs utilize samples much more efficiently than DNNs.

\subsection{Generalization to Larger Area}

Apart from the ability to contend with heterogeneity, another major advantage of HIGNN is the scalability to larger problem scales. Next, we show that HIGNN is capable of handling networks with increasing number of links after trained on small network instances of fixed size. In contrast, DNNs are unable to deal with wireless networks larger than the setting of training set \textit{i.e.}, 8 SISO links and 4 MISO links. As DNNs are not scalable to larger networks, we do not include DNNs in the following results.

\begin{figure}[htbp]
\centerline{\includegraphics[width=3.5in]{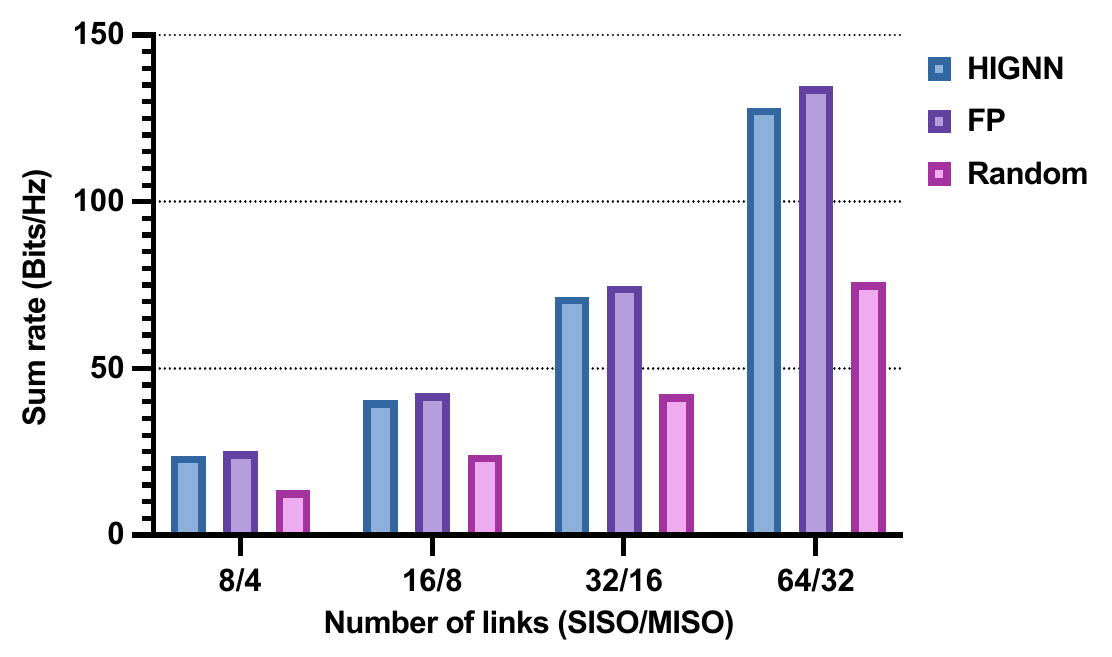}}
\caption{Generalization performance of HIGNN to networks with expanding area.}
\label{larger_area}
\end{figure}

We fix the density of links and enlarge the problem scale by adding up the area length to hold more links. Starting from 8 SISO and 4 MISO links in a $400\times400{\rm m}^{2}$ area, each time we double the number of links by augmenting the area length by $\sqrt{2}$. As shown in Fig. \ref{larger_area}, the relative performance of HIGNN against FP still remains above $95\%$ when the network size is increased by 8 times.

\subsection{Generalization to Higher Link Density}

\begin{figure}[htbp]
\centerline{\includegraphics[width=3.5in]{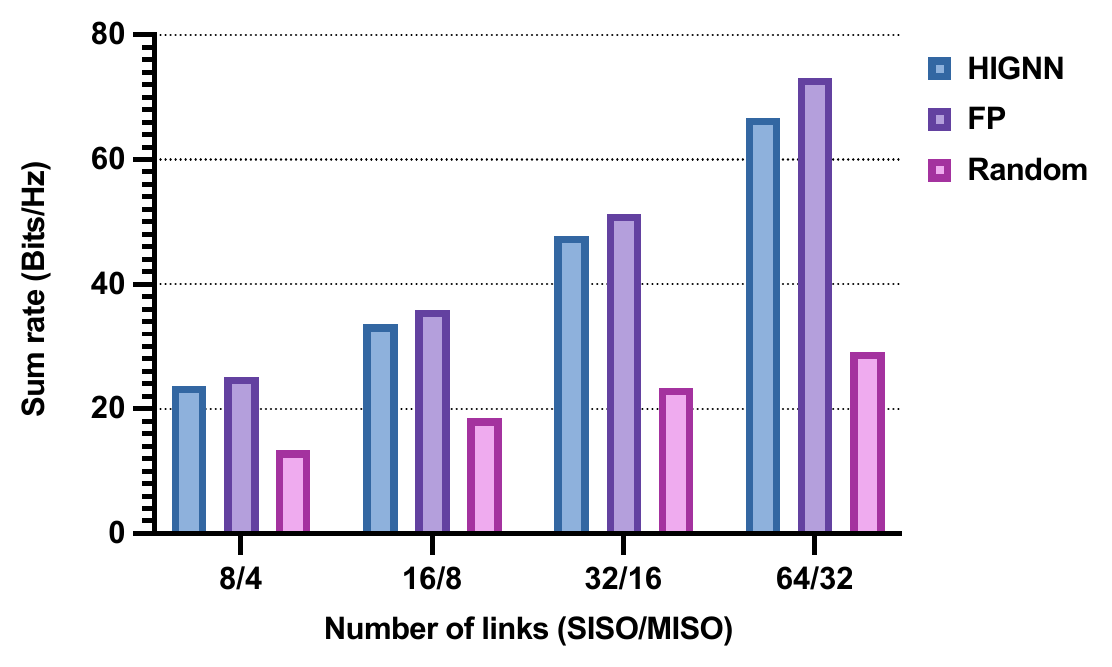}}
\caption{Generalization performance of HIGNN to networks with higher link density.}
\label{higher_density}
\end{figure}

Next, we let the trained HIGNN carry out a more challenging task. With the area fixed, statistical distribution of interference can be altered by the growth in link density. Beginning at 8 SISO and 4 MISO links in $400\times400{\rm m}^{2}$, each time we double the number of links within the same area. Results are displayed in Fig. \ref{higher_density}.

Compared with the results in Fig. \ref{larger_area}, the per-link performance of both HIGNN and FP deteriorates due to more severe interference in congested settings. The relative performance of HIGNN against FP is $93.83\%$, $92.99\%$ and $91.16\%$ when the number of links increases to $24$, $48$ and $96$, respectively. Performance gap is enlarged with the increasing link density. Nonetheless, the trend of performance degradation is rather slow. The relative performance of HIGNN is kept above $90\%$ when the link density is 8 times the amount during training. HIGNN is shown to hold strong transference ability when the statistical characteristics of test data deviate from training data. 

\subsection{Execution Time and Complexity}

One significant improvement of ML-based methods on iterative algorithms like FP is the reduction in execution time while keeping strong performance. Here, we run both HIGNN and FP on the same hardware and no GPU acceleration is used by HIGNN for fair competition. For each setting, we randomly generate 100 channel instances and calculate the average execution time. 

\begin{figure}[htbp]
\centerline{\includegraphics[width=3.5in]{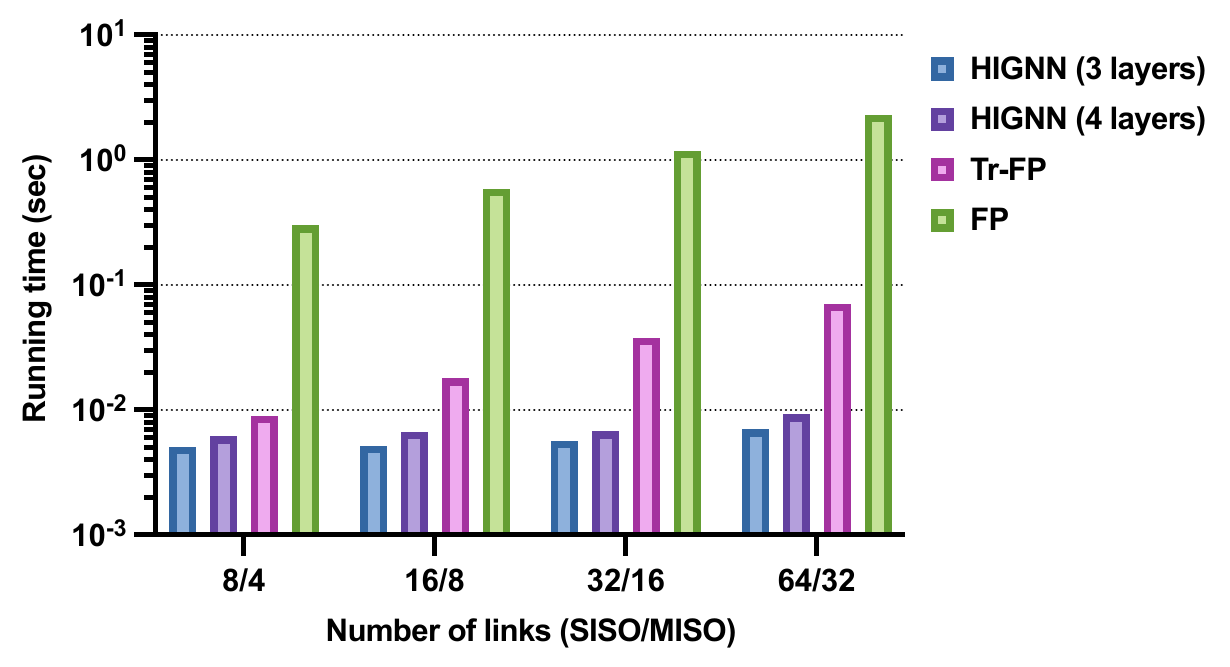}}
\caption{Execution time (in logarithm) of closed-form FP and HIGNN. Note that running time is shown in logarithm.}
\label{running_time}
\end{figure}

The average execution time to solve problems with different size is shown in Fig. \ref{running_time}. Here, we also calculate the running time of 3 iterations in FP (referred as Tr-FP) for clearer comparison. The running time of FP grows dramatically with the problem size, while the increase for HIGNN is insignificant. Compared with FP, acceleration of over 300 times is achieved by HIGNN when the problem scales to 96 links. 

In terms of time complexity, both HIGNN and FP have the per-iteration complexity of $\mathcal{O}(K^{2})$. Nevertheless, in beamforming design, bisection search is executed in each iteration of FP to find the local optimum, while HIGNN only requires forward computation. HIGNN generally shows much higher execution efficiency than FP.

\section{Conclusion}
This article extends the homogeneous GNN-based RA framework to heterogeneous settings, which could handle interference in more complicated but realistic WCNs. After the graph modeling of heterogeneous interference relations, we show that HIGNN is capable of learning RA policies from scratch. Future work will explore the optimal solution structures of more complex RA problems and incorporate them to HIGNN to further improve the learning efficiency.

\bibliography{bib}{}
\bibliographystyle{IEEEtran}

\end{document}